\documentclass[10pt,twocolumn,letterpaper]{article}

\usepackage{iccv}
\usepackage{times}
\usepackage{epsfig}
\usepackage{graphicx}
\usepackage{amsmath}
\usepackage{amssymb}
\usepackage{url}
\usepackage{xcolor}
\usepackage{tabularx}
\usepackage{booktabs}
\usepackage{adjustbox}

\usepackage[breaklinks=true,bookmarks=false]{hyperref}

\iccvfinalcopy 

\def\iccvPaperID{****} 

\begin{document}

\title{A Classification approach towards Unsupervised Learning of Visual Representations}

\author{Aditya Vora\\
Johnson Controls Inc.\\
{\tt\small voraaditya898@gmail.com}
}

\maketitle

\begin{abstract}
In this paper, we present a technique for unsupervised learning of visual representations. Specifically, we train a model for foreground and background classification task, in the process of which it learns visual representations. Foreground and background patches for training come after mining for such patches from hundreds and thousands of unlabelled videos available on the web which we extract using a proposed patch extraction algorithm. Without using any supervision, with just using $150,000$ unlabelled videos and the PASCAL VOC 2007 dataset, we train a object recognition model that achieves $45.3$ mAP which is close to the best performing unsupervised feature learning technique whereas better than many other proposed algorithms. The code for patch extraction is implemented in Matlab and available open source at the following \href{https://github.com/aditya-vora/unsupervised_patch_extraction/}{link} . 
\end{abstract}

\section{Introduction}

ConvNet is a type of deep learning model which have proved to be an effective tool for learning generalized visual representations of various entities appearing in the image \cite{he2016deep, long2015fully, ren2015faster, he2017mask}. The basic approach towards learning features for a specific task is to start of by a base image classification model \cite{krizhevsky2012imagenet, he2016deep} where the features are learned with a fully supervised approach using large scale datasets like ImageNet \cite{russakovsky2015imagenet}, and then perform transfer learning on this base model for the new desired task with a new dataset. This approach of ``building up on base features" is considered to be a standard approach for learning good features, specific to a particular task. However, there are a few drawbacks in this approach: 1) It requires huge amount of labelled training data in order to learn the base classification model, which eventually takes a lot of human efforts. 2) This kind of feature learning strategy lacks some sort \textit{resemblance} with how the human/animal learns object features about various objects it sees around the world. Human visual system makes use of the massive amount of unlabelled data that it observes in the world in order to learn about the objects it observes. Thus, it becomes extremely beneficial if we can learn good features by leveraging the massive amount of unlabelled data present on the internet.

\begin{figure}[!t]
\centering
\includegraphics[scale=.3]{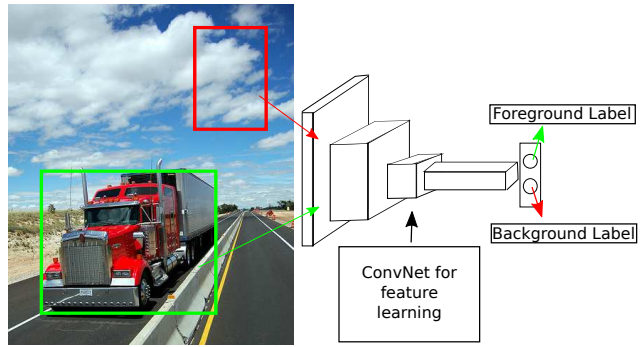}
\caption{From the initial set of video frames we extract foreground and background patches (shown in green and red boxes respectively) from the frames using a proposed approach. We then train a CNN to predict the labels of these foreground and background patches using a binary classification model. After training this model, we transfer the learned features to a new task like object recognition.}
\end{figure}

People have tried to solve the problem of unsupervised feature learning by targeting the problem from several angles, \cite{gao2016object} and \cite{wang2015unsupervised} have come up with an approach where they try to learn a feature embedding where the distance between two object patches would be less, whereas the distance between the object and non object will be more. However, both the approaches have their own drawbacks, 1) \cite{wang2015unsupervised} use tracking in order to sample \textit{dynamic} object in a scene and consider the patch sampled from the first frame and the last frame as the same object and use that pair of samples during the training of the Siamese network. However, such an approach is biased towards the dynamic object in the scene and neglects the static objects, because of which it creates some noisy samples during training (as there is a high probability that the non-object patches is sampled from the regions containing the static objects) because of which quality of training sample patches are affected. Moreover, applying tracking to process massive amount of unlabelled video is computationally expensive, and error in tracking may influence the quality of the patches used for feature learning. 2) \cite{gao2016object} made use of the temporal coherence of the same object in a video as a free supervision in order to learn features. The patches extracted from adjacent frames in the video represents same object in the embedded space. The overall approach was same as \cite{wang2015unsupervised} but the only difference was that, instead of extracting patches with tracking, they made use of object proposals with high IoU in adjacent frames as the proposals representing the same object. However, considering the objects only in the adjacent frames, the algorithm fails to consider the pose variations that the object may undergo throughout the video because of which the training process will not get variety of training samples. \cite{pathak2017learning} came up with a segmentation based approach where they perform an unsupervised motion segmentation on all the videos and then use that segments for learning features. However, such an approach is biased by the accuracy of the motion segmentation algorithm and thus directly affect the feature learning process.

By considering these limitations, we propose a new approach towards feature learning from unlabelled videos. We target the feature learning process through a foreground/background classification approach. Training data for the classification task comes from a foreground/background patch extraction technique proposed by us. Our approach for unsupervised feature learning tries to exploit the fundamental feature difference between the foreground and the background region in an image through a machine learning model. Our approach for foreground and background patch extraction is based on the fact that in a video the foreground region is generally the salient and dynamic object in the scene, whereas the background region is the least salient and the region which is nearly static.

\section{Related Work}

Previously many contributions were been made in this area of unsupervised feature learning. A lot of unsupervised learning algorithms have been proposed which work towards learning good feature representations directly from the unlabelled raw data. Some of these are autoencoders (\cite{hinton2006reducing, bengio2009learning}). By reducing the number of parameters in the hidden layers hierarchically, it forces the network to learn low level representations of an image which has a least square approximation to the original image. As a result of this, the network learns to retain the important features required to build the entire image. However, these techniques have some drawbacks: 1) Just learning the lower dimension representation of an image does not help in learning good visual representations. 2) This process does not provide any ques to the network about what can be good feature extraction areas in an image. Because of this autoencoders does not give good results when we transfer the features to a new task.

So instead of learning features using completely unsupervised approaches, a significant amount of work has been done to figure out alternative forms of supervision that can be provided to the learning algorithm (except labels) so that the feature learning process becomes significantly easy for the neural network compared to pure unsupervised approaches discussed earlier. For instance \cite{pathak2017learning} had trained a segmentation model for foreground/background segmentation, but instead of using the manually annotated segments as ground-truth they estimated the labels for all the frames of the video with an unsupervised motion segmentation algorithm  (\cite{faktor2014video}) and used that estimated segments of the video frames as the corresponding ground-truth. \cite{doersch2015unsupervised} trained a CNN in order to predict the relative location of a second patch in a pair compared to the first. \cite{wang2015unsupervised} came up with an approach where they try to localize a dynamic object in a scene using dense trajectories (\cite{wang2013action}). In order to learn a good feature representation, they learn a feature embedding in a low dimensional space where the distance between the patches representing the same object in a scene is less and that representing different objects is more. \cite{gao2016object} is a similar approach to \cite{wang2015unsupervised} however instead of relying on tracking to get a more diverse poses of same object, they focus on the idea that two spatio-temporally close region proposals should be embedded close in the deep feature space. \cite{noroozi2016unsupervised} tries to learn feature representations by solving a jigsaw puzzle on a $3\times3$ grid by arranging shuffled patches cropped from the grid.

\section{Approach}

Given a large set of unlabelled videos, our task is to learn visual representations in a completely unsupervised fashion. Our approach is a two stage framework: 1) Using a proposed technique, extract foreground and background patches from video frames. 2) Train a classifier for foreground/background classification using the patches extracted from the video frames, which act as supervisory signals to our model during training.

\subsection{Extracting FG and BG patches}
We start by downloading the YFCC100m \cite{thomee2016yfcc100m} video dataset. It consists total of around $700,000$ videos. However, we do not use this version of dataset, rather we use the pruned down version of the dataset provided by \cite{pathak2017learning}. It contains around $205,000$ videos, where each video consists of around $5-10$ frames sampled randomly from the original videos. From these set of frames we extract our foreground and background patches which are further used for feature learning process. 

As a pre-processing step, we try to remove all the set of frames that contain any scene cuts/transitions. In order to do so we take two factors into consideration: 1) Pixel space correlation, where low correlation between frame pairs usually corresponds to scene cuts. We select those frames that have correlation greater than $0.1$. 2) Average Intensity, where we select those frames that satisfy upper and lower average intensity limit. The limits are set to $(50,200)$.

\begin{figure*}[!t]
\centering
\includegraphics[scale=.2]{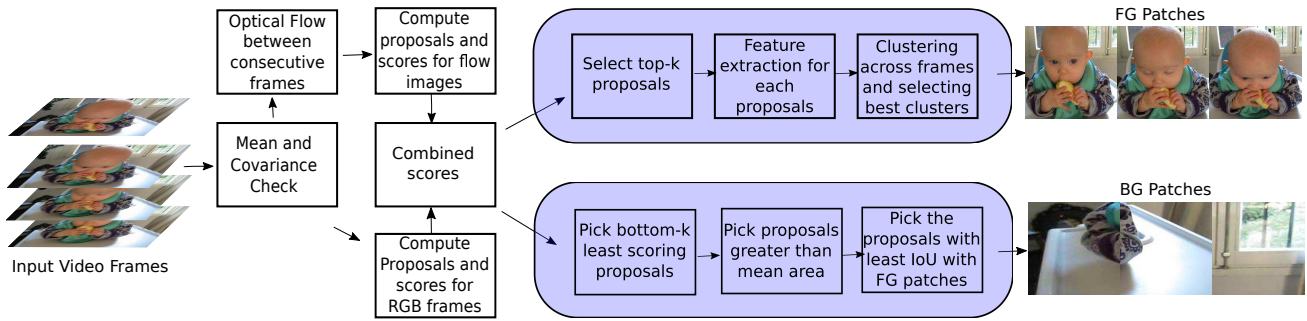}
\caption{Overall pipeline of extracting the foreground and background patches from the video. From the original set of frames we filter out outlier frames based on the the pixel correlation and the average intensity of the frames. Then the remaining set of frames undergoes the foreground background patch extraction process with our proposed approach.}
\label{fig: overall pipeline}
\end{figure*}

After filtering out the outlier frames based on the pixel correlation and average intensity, we then perform the patch extraction process i.e. from each frame of the video we extract a high probable foreground as well as a background patch. Our patch extraction algorithm is based on the fact that a high probable foreground region in a video is generally a region with high motion as well as salient score whereas opposite holds true for high probable background patches. To start with, let us say that for a video $V$, after the filtering step we are left with $F=\{F_{1}, F_{2},...,F_{N}\}$ set of frames. As mentioned earlier in order to make use of the motion and salient cues for the selection of the foreground and background patches, we compute optical flow $O=\{O_{1}, O_{2},...,O_{N-1}\}$ between each frame of the video using \cite{liu2009beyond}. We then compute object proposals using a technique called EdgeBoxes proposed by \cite{ZitnickECCV14edgeBoxes}. This object proposal algorithm scores each proposal based on the edge contour that are wholly-contained in it, which indicates that the proposed region contains a whole object. We make use of the these scores in order to get the appearance score, $s_{a}$ and motion scores, $s_{m}$ of each proposed region. Specifically, we extract $\sim500$ proposals each from the RGB frame belonging to the set $F$ as well as from the corresponding optical flow image from the set $O$. Here, optical flow image is obtained by computing the optical flow magnitude from the corresponding $X$ and $Y$ flow component pixel locations. Thus we get a total of $\sim1000$ proposals per frame. However, we need to ensure that each proposal of these $1000$ proposals have both appearance score, $s_{a}$ as well as motion score $s_{m}$. As the first $500$ proposals comes from the RGB frame the score corresponding to the proposals computed by the algorithm is the appearance score $s_{a}$, whereas in case of the rest of the $500$ proposals which have come from the optical flow image the scores computed by the algorithm corresponds to motion score $s_{m}$. However the proposals from the RGB frame donot have motion score whereas the proposals from the optical flow image do not have the appearance scores. Thus in order to compute both scores for each proposal, we take the corresponding object proposal region of the RGB image on the optical flow image and compute it's scores, which will give us the motion score of the object proposals computed from the RGB frames. In the same way we take the object proposal region of the optical flow image on the RGB image and compute it's scores, which will give us the appearance scores of the proposals computed from the optical flow image. Doing this we obtain both appearance score, $s_{a}$ and motion score, $s_{m}$ for each of the $\sim1000$ computed proposals. We then compute the product of $s_{a}\times s_{m}$ after normalizing the scores in order to get the overall objectness score $s$, and retain top $15$ high scoring proposals per frame.

After the previous step, we have retained top-$15$ proposals per frame which have the highest overall objectness score, $s$. In order to ensure that variety of foreground patches containing object with different poses are provided to the model during the training process, we look for the regions that are spread out in time. We do a nearest neighbour search to cluster the proposals across all the frames and select the one with maximum score. Specifically, we consider the object proposals from the first frame as the seed proposals $R$. Then for each seed proposal in $R$ we compute the feature similarity between all the object proposals in the subsequent frames. In order to do this we use $L_{2}$ normalized $2048$ D feature vector which is obtained after removing the final classification layer from a 50-layer ResNet model \cite{he2016deep}. For this purpose we make use of the publicly available pre-trained ResNet model and we do a forward pass of each object proposal through the network to obtain the feature representation. Thus, we compute the feature similarity by taking the inner product between the proposals in the seed set $R$ and all the remaining proposals in the rest of the frames and select the matching proposal with the seed as the proposal that has maximum inner product score among the proposals in that corresponding frame. After computing the matching proposals with the seed set throughout all the frames, we than assign a cluster score to all the selected clusters and select the one with maximum score. The cluster score is computed as a cummulative score of the overall score of the proposal $s$, and the inner product score of that proposal with the seed proposal. The cluster scoring equation is shown in Equation \ref{eq:1}.

\begin{equation}
s(c) = \sum_{j} s(p^{j}) * (\phi(p^{j})^{T}\phi(p^{seed}))
\label{eq:1}
\end{equation}

Here $s(c)$ is the cluster score corresponding to a set of proposals, $s(p^{j})$ is the overall score of the $j$th object proposal. $\phi(.)$ represents the $L_{2}$ normalized deep feature representation of the corresponding proposal. We select the cluster with the maximum score and thus the proposal members of that corresponding cluster will be considered as the foreground patches. The selected cluster will have two properties, first it will have high overall score, thus the proposals contained in this cluster will have high overall scoring regions. Second, these proposals are spread out in time thus satifying the criteria of having variety of object poses for training. 

\begin{figure*}[!t]
\centering
\includegraphics[scale=.165]{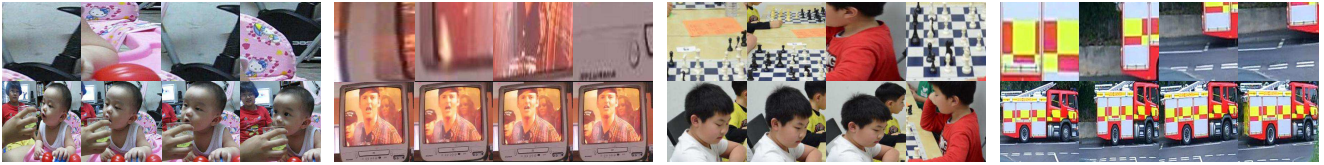}
\caption{Some foreground (bottom) and background (top) patches extracted by our proposed algorithm.}
\label{fig: fg/bg patch sample}
\end{figure*}

After extracting the foreground region from the selected set of frames we now extract the background regions. However, in order to have a good set of regions representing the background, we make use of the appearance and the motion scores (i.e. overall score $s=s_{a}\times s_{m}$). We extract background regions using the simple fact that regions with lower appearance and motion score will represent the background. This is true from the basic observation that, objects that are moving in a scene and have a high appearance score are mainly the foreground region in a video (Dynamic Object). However other than this, there are also objects in the video that are static. However, for such regions the motion scores will be less but the appearances scores would be more. Thus, only the background regions (i.e. which has no object and motion) have less appearance as well as motion scores. Thus we follow this strategy in order to extract background patches from the video frames. We consider $100$ proposals having least overall score $s$ to be considered as a background region. But doing this we face a small issue. The proposals having the least scores are normally very small regions. In order to make sure that patches with considerable area are taken into consideration, we choose proposals that have area greater than the mean area of all the $100$ proposals. Moreover, we also want to ensure that the selected background regions do not overlap with the selected foreground patch. In order to this, we compute the IOU of the remaining object proposals after filtering based on area, with the foreground patch of the corresponding frame, and select the object proposal region that has the least bounding box overlap with the foreground patch. In this manner we can get regions that are big in area, as well as quite distinct from the foreground region because of which the quality of the background regions are good. The overall pipeline for foreground and background patch extraction in explained in Figure \ref{fig: overall pipeline}. Results of the foreground and background patches on a few set of videos is shown in Figure \ref{fig: fg/bg patch sample}. It can be seen that our algorithm is capable of extracting discriminative patches from any video. The top row of patches are the background patches whereas the bottom row of patches are the foreground patches. As seen in the figure background patches comes from highly static and non-salient regions of an image because of which they are sort of random and does not convey any information and thus forms a good samples representing background. Also, the foreground patches as it can be seen from the figure, contains mostly the dynamic as well as the salient object in the scene because of which most of the time the foreground patch contains some object in the scene. Giving these patches for training to the classifier, will help the classification model to learn how to discriminate between a general notion of foreground (object) and background (non-object). 

\section{Training of FG/BG Classifier}
After the extraction of appropriate number of foreground and background patches using our proposed approach, we then train our FG/BG classifier using the extracted patches for training. In order to have a consistent comparison with previous works, we use AlexNet \cite{krizhevsky2012imagenet} as our base model. We change the number of output nodes from $1000$ which is the number of classes used in the classification of ImageNet \cite{deng2009imagenet} dataset, to $2$ (Foreground and Background). The input dimension to the CNN is fixed to $227 \times 227$ which is the standard input to the AlexNet architecture. We train the model for approximately $600,000$ iterations with each iteration having a batch size of $32$. The input undergoes a standard normalization of mean subtraction and scaling. The learning rate while training the base model is set to $0.001$ and the entire model is trained with stochastic gradient descent. We keep these hyper-parameters settings same throughout our experiments i.e. for training all the versions of the base models.

\section{Experiments}

\begin{table*}
\begin{center}
\begin{adjustbox}{max width=\textwidth}
\begin{tabular}{|l|c|c|c|c|c|c|c|c|c|c|c|c|c|c|c|c|c|c|c|c|c|c|}
\hline
\textbf{Amount of Data} & aero & bicycle & bird & boat & bottle & bus  & car  & cat  & chair & cow  & diningtable & dog  & horse & motorbike & person & pottedplant & sheep & sofa & train & tvmonitor & \textbf{mAP} \\
\hline\hline
\textbf{50,000}            & 53.3 & 51.2    & 28.4 & 26.4 & 10.2   & 52.4 & 58.5 & 44.1 & 17.1  & 38.8 & 45.2        & 35.6 & 60.9  & 56.8      & 43.4   & 16.7        & 30.4  & 42.5 & 56.1  & 44.2      & 40.7         \\
\textbf{100,000}           & 55.8 & 53.7    & 28.5 & 27.1 & 13.2   & 57.8 & 60.4 & 44.6 & 18.3  & 40.1 & 47.3        & 36.9 & 61.3  & 55.9      & 45.7   & 17.9        & 30.5  & 42.3 & 58.2  & 46.8      & 42.1         \\
\textbf{150,000}           & 58.1 & 56.7    & 32.3 & 28.2 & 17.6   & 60.6 & 63.3 & 47.6 & 20.7  & 43.9 & 50.6        & 40.8 & 65.2  & 58.1      & 49.7   & 22.9        & 35.4  & 44.5 & 60.8  & 49.3      & 45.3         \\
\hline
\end{tabular}
\end{adjustbox}
\end{center}

\caption{Performance with varying the amount of training data for the classification model. As we can observe that there is a trend of increasing mAP with increasing the amount of training data for the classification model. Not only average mAP what we observe is that there is a increasing trend of mAP across object across all the classes available in the dataset.}

\label{tab:2}

\end{table*}

\begin{table}
\begin{center}
\begin{tabular}{|l|c|}
\hline
\textbf{Method} & \textbf{mAP} \\
\hline\hline
Gao et al. (2016) \cite{gao2016object} & 46 \\
Doersch et al. (2015) \cite{doersch2015unsupervised}  & 47.8 \\
Pathak et al. (2017) \cite{pathak2017learning}  & 48.6 \\
Wang and Gupta (2015) \cite{wang2015unsupervised} & 43.5 \\
Agrawal et al. (2015) \cite{agrawal2015learning} & 37.4          \\
Pathak et al. (2016) \cite{pathak2016context}  & 39.1          \\
Owens et al. (2016) \cite{owens2016ambient}   & 42.9          \\
Donahue et al. (2016) \cite{donahue2016adversarial} & 42.8          \\\hline
\textbf{Ours}         & \textbf{45.3} \\
\hline
\end{tabular}
\end{center}

\caption{Our results on test split of PASCAL VOC 2007 \cite{pascal-voc-2007} dataset compared to other techniques. This results that we generate are using 150,000 videos from the YFCC100m dataset \cite{thomee2016yfcc100m}.}

\label{tab1: map results}

\end{table}

\subsection{Evaluating the learned representation}
The quality of the learned representations are determined from how are the results when the features are transferred to a new task. In our case we transfer the learned features from our base model to a new task of object recognition. Concretely, we take our trained foreground and background classification model as a base model in fast-RCNN \cite{girshick2015fast} pipeline, which performs the object recognition task and then train the entire object recognition model with the trainval split of PASCAL VOC 2007 dataset \cite{pascal-voc-2007}. The trainval split of the PASCAL VOC 2007 dataset has $5011$ images which are available for training the object recognition model. During the training the learning rate is $0.005$ which is kept constant throughout the experimentation. The trained model is then tested on the test split of the of the corresponding VOC 2007 dataset which has around $4952$ test images.

Table \ref{tab1: map results} shows the results obtained by transfer learning on the object recognition task. As mentioned above these results are on the test split of the VOC 2007 dataset. Moreover, these results are obtained after training the base classification model with $150,000$ videos from the YFCC100m dataset i.e. foreground and background patches are extracted from $150,000$ videos and then those patches are provided to the model during the training process. We compare our mAP results with other previous unsupervised feature learning techniques \cite{doersch2015unsupervised, pathak2017learning, gao2016object, wang2015unsupervised, agrawal2015learning, pathak2016context, owens2016ambient, donahue2016adversarial}. As it can be seen from the table our base model in the object recognition pipeline achieves a mAP of $45.3$. However, \cite{pathak2017learning} achieve best mAP accuracy which is $48.6$ among all the unsupervised learning techniques. However, our algorithm performs better than many unsupervised techniques like \cite{wang2015unsupervised, agrawal2015learning, pathak2016context, owens2016ambient, donahue2016adversarial}.

\subsection{Performance with amount of training data}
One quality of deep learning algorithms is that, their performance improves with data. We experimented, how the quality of visual representations changes with the amount of data. In order to do this we trained the classification models with varying amount of data and then check whether the accuracy of the final transfer learning task is improved with the increase in the amount of data. We experimented by partitioning the total data in a chunks of $50,000$ videos. We train our foreground/background classification model with $50,000$, $100,000$, and $150,000$ videos. After training the classification model we then use that model as the base network in the fast-RCNN task just as we did previously. As it can seen from Table. \ref{tab:2} by using only $150,000$ videos from the video dataset for training the classification model we are already better than many other unsupervised feature learning techniques \cite{wang2015unsupervised, agrawal2015learning, pathak2016context, owens2016ambient, donahue2016adversarial}. Moreover this number of $45.3$ mAP is further expected to improve if we make use of all the training data i.e. $200,000$ videos. 

\section{Conclusion}
We have presented a simple approach to unsupervised feature learning by using foreground/background classification. We also have proposed a patch extraction algorithm which will help in extracting the foreground and background patches which will act as supervisory signal in the feature learning process. We noted that our network performs better than many other unsupervised feature learning techniques however being comparatively simpler approach. Moreover we also observed that the quality of the learned representations improve with the amount of training data.

{\small
\bibliographystyle{ieee}
\bibliography{egbib}
}

\end{document}